\documentclass{article}

\usepackage{arxiv}

\usepackage{cite}
\usepackage{float}
\usepackage{amsmath}
\usepackage{listings}
\usepackage{url}
\usepackage{graphicx}
\usepackage{amssymb}
\usepackage{bm}
\usepackage{comment}
\usepackage{authblk}
\usepackage{multirow}
\usepackage{amsthm}
\usepackage{caption}
\usepackage{graphicx}
\usepackage{algorithm}
\usepackage{algorithmic}
\usepackage{url} 
\usepackage{subcaption}

\usepackage{xcolor}
\usepackage{algorithm}
\usepackage{algorithmic}
\usepackage{hyperref}
\usepackage{cite}
\date{}                                           
\def\U{{\mathcal{U}}}
\def\X{{\mathcal{X}}}
\def\A{{\mathcal{A}}}
\newcommand{\size}[1]{\in\mathbb{R}^{#1}}

\newcommand{\abs}[1]{\left\lvert#1\right\rvert}
\newcommand{\norm}[1]{\lVert#1\rVert}
\newcommand{\fnorm}[1]{\lVert#1\rVert_{\mathrm{F}}^{2}}
\newcommand{\tran}[1]{#1^{\mathrm{T}}}
\newcommand{\tranh}[1]{#1^{\mathrm{H}}}
\newcommand{\inv}[1]{#1^{\dagger}}
\newcommand{\bibra}[1]{[\![#1]\!]}
\newcommand{\st}{{\text{s.t.}}}
\newcommand\tqh[1]{\textcolor{orange}{#1}}
\newcommand\tcc[1]{\textcolor{teal}{#1}}
\def\x{{\mathbf x}}
\def\L{{\cal L}}

\title{Tensorized LSSVMs for Multitask  Regression
}
\author{Jiani~Liu{$^1$}, Qinghua~Tao$^2$, Ce Zhu{$^{1}$}\thanks{Corresponding Author.  This research is partially supported by the National Natural Science Foundation of China (NSFC) under Grant U19A2052, Grant 62020106011 and Grant 62171088. Johan A.K. Suykens and Qinghua Tao acknowledge the supports from iBOF project Tensor Tools for Taming the Curse (3E221427), European Research Council (ERC) Advanced Grant E-DUALITY (787960), KU Leuven 
Grant CoE PFV/10/002, Grant FWO GOA4917N, EU H2020 
ICT-48 Network TAILOR, and Leuven.AI Institute.\\ Email: jianiliu@std.uestc.edu.cn,~qinghua.tao@esat.kuleuven.be,~eczhu@uestc.edu.cn,~yipengliu@uestc.edu.cn,~johan.suykens@esat.kuleuven.be}~, Yipeng Liu{$^1$}, Johan A.K. Suykens $^2$\\ 
$^1$ School of Information and Communication Engineering, University of Electronic Science and Technology of China, 610054 Chengdu, China \\
$^2$ ESAT-STADIUS, KU Leuven, 3001 Heverlee, Belgium

}

\begin{document}
\maketitle

\begin{abstract}
    Multitask learning (MTL) can utilize the relatedness between multiple tasks for performance improvement. The advent of multimodal data allows tasks to be referenced by multiple indices. High-order tensors are capable of providing efficient representations for such tasks, while preserving structural task-relations. In this paper, a new MTL method is proposed by leveraging low-rank tensor analysis and constructing tensorized Least Squares Support Vector Machines, namely the tLSSVM-MTL, where multilinear modelling and its nonlinear extensions can be flexibly exerted. We employ a high-order tensor for all the weights with each mode relating to an index and factorize it with CP decomposition, assigning a shared factor for all tasks and retaining task-specific latent factors along each index. Then an alternating algorithm is derived for the nonconvex optimization, where each resulting subproblem is solved by a linear system. Experimental results demonstrate promising performances of our tLSSVM-MTL.
\end{abstract}


\section{Introduction}
\label{sec:intro}

Multitask learning (MTL) lies on the exploitation of the coupling information across different tasks, so as to benefit the parameter estimation for each individual task \cite{caruana1997multitask,argyriou2008convex,xu2011multi}.  MTL has been widely applied in many fields, such as social sciences\cite{xu2014multi, Evgeniou2004regu, Evgeniou2005learn}, medical diagnosis \cite{han2012implementation, Liang2009Predictive}, etc. Various MTL methods have been developed and shown promising performance for related tasks. Among them, support vector machines (SVMs) 
get great success \cite{cortes1995support}. Specifically, based on the minimization of regularization functionals,  the regularized MTL is proposed in \cite{Evgeniou2004regu, Evgeniou2005learn} with kernels including a task–coupling parameter. An MTL method based on SVM+, as an extension of SVM, is developed in \cite{lichen2008connection} and compared with standard SVMs in \cite{Liang2009Predictive} and regularized MTL in \cite{han2012implementation}. Moreover, the least squares SVM (LSSVM) \cite{suykens1999least} is  also generalized  for MTL \cite{xu2014multi}, where the inequality constraints in SVMs are modified into equality ones and a linear system is solved in dual instead of the typical quadratic programming. These SVM-based MTL methods were all applied with the typical vector/matrix expressions.

Tensors,  a natural extension for vectors and matrices, provide a more effective way to preserve multimodal information and describe complex  dependencies\cite{liu2022tensor,liu2021tensor}. Different usages of tensor representations have been successfully applied to MTL \cite{zheng2019multitask,xu2019spatio,zhang2019tensor,yang2017deep,zhang2020deep,romera2013multilinear,wimalawarne2014multitask,zhao2019multilinear}.  For instance, motivated by the multidimensional input, \cite{zheng2019multitask} proposed to factorize the weight tensor for each task into a sparse task-specific part and a low rank shared part. In \cite{xu2019spatio}, it formulates the input as a tensor and extracts its spatial and temporal latent factors, based on which a prediction model is built. It is also intriguing to  encode the  projection matrices of all classifiers into tensors and apply tensor nuclear norm constraints for task relations \cite{zhang2019tensor,yang2017deep}.  


The aforementioned works are all set with a single index for the involved tasks. In practice,  tasks can be referenced by multiple indices with physical meanings. Taking a multimodal data task for example, restaurant recommendations consider different aspects of rating (e.g., food and service) and customers. It naturally leads to $ T_1 \times T_2 $ tasks spanned by two indices, and thus a single index fails to preserve such information. Therefore, \cite{romera2013multilinear} considered tasks with multiple indices and imposed low Tucker rank regularization over the stacked coefficient tensor {to explore task relations.
In \cite{romera2013multilinear}, the applied Tucker decomposition
can suffer from a dimensionality curse if the tensor order increases.   For rank minimization, a convex relaxation is used to
handle the whole weight tensor in each iteration and thereby can be problematic for large-scale data. Two variants were later developed in \cite{wimalawarne2014multitask,zhao2019multilinear} with different convex relaxations for Tucker rank minimization.  Though nonconvex optimization was also considered in  \cite{romera2013multilinear}, it required adjusting several ranks within Tucker, making the tuning procedures  rather complicated. Besides, they all considered multilinear modelling, while nonlinearity is highly desirable for well describing complex data  and tasks.} 


In this paper, we develop a tensorized MTL method for regression by leveraging  LSSVMs, namely the tLSSVM-MTL, {which constructs a high-order weight tensor on LSSVMs and indexes the tasks along different modes into groups by multiple indices.} {Unlike \cite{romera2013multilinear,wimalawarne2014multitask,zhao2019multilinear}, we factorize the constructed tensor into CP forms {since the factors are easy to explain from subspace perspective}, and} enable all tasks to share a common latent factor and meanwhile retain task-specific factors. 
 In our method, both linear and nonlinear feature maps (or kernels) can be flexibly employed.  For optimization,  an alternating minimization strategy is proposed with each subproblem solved by a linear system in the dual. 
 Numerical experiments show  advantageous performances of our tLSSVM-MTL over matrix-based and existing tensor-based MTL methods.

The next section  gives some  premieres. Section \ref{sec: method} presents the  modelling and optimization for our tLSSVM-MTL. Experimental results and conclusions are  in Sections \ref{sec: experiment} and \ref{sec: conclusion}. 

\section{Preliminaries}
\label{sec:preliminarie}
Scalars, vectors, matrices, and tensors are represented as $x$, $\mathbf{x}$, $\mathbf{X}$, and $\mathcal{X}$, respectively. 
For clarity, we denote the row and the column in a matrix $\mathbf{X}$ as $\mathbf{X}[i,:]^T = \mathbf{x}_{i,:}$ and $\mathbf{X}[:, j] = \mathbf{x}_{:,j}$. 

\vspace{0.1cm}
\noindent\textbf{CP decomposition }\cite{carroll1970analysis,phan2021canonical} Given a tensor $\mathcal{X}\in\mathbb{R}^{I_1\times \cdots \times I_N}$, CP decomposition
factorizes the tensor into a summation of several rank-one components as 
$\mathcal{X}=\sum_{k=1}^{K} \mathbf{u}^{1}_k \circ \cdots \circ \mathbf{u}^{N}_k,$
where $K$ is the  CP rank indicating the 
 smallest number of rank-one components required in this representation.
We represent the CP decomposition as $\mathcal{X}=[\![\mathbf{U}^{1},\ldots,\mathbf{U}^{N} ]\!]$  with $\mathbf{U}^{n}=[\mathbf{u}^{n}_1,\ldots,\mathbf{u}^{n}_K]$ for $n=1,\ldots,N$. 

\vspace{0.1cm}
\noindent{\textbf{LSSVM}}
LSSVM \cite{suykens1999least} is a variant of SVMs \cite{cortes1995support} by forming equality constraints.
For regression with data $\{\mathbf{x}_i,y_i\}_{i=1}^m$, the primal problem of
LSSVM is given as:
\begin{eqnarray}
\underset{\mathbf{w},  b, \mathbf{e}}{\min } & J\left(\mathbf{w},  b, \mathbf{e}\right)=\frac{{C}}{2} \sum\nolimits_{i=1}^{m} \left(e_i\right)^{2}+\frac{1}{2} \mathbf{w}^{\top}\mathbf{w} \nonumber\\ 
\text { s.~t.~~ } &  \mathbf{w}^{\top} \phi(\mathbf{x}_{i})+b=y_i-e_{i},  \nonumber
\end{eqnarray}
where {$\phi: \mathbb{R}^{d} \mapsto \mathbb{R}^{d_h}$  is the feature mapping function,  $\mathbf{w} \in \mathbb R^{d_h}$  and $b\in \mathbb R$ are the  modelling coefficients},  $e_i$ denotes the point-wise regression error, and $C>0$ is the regularization hyperparameter. In LSSVMs, the Lagrangian dual problem gives  a linear system, instead of the quadratic programming  in classic SVMs, making certain problems more  tractable.
           


\section{Tensorized LSSVMs for MTL}
\label{sec: method}
\subsection{Tensorized Modelling}\label{sec:w:tensor}
\begin{figure}[t]
    \centering
\includegraphics[scale=1.5]{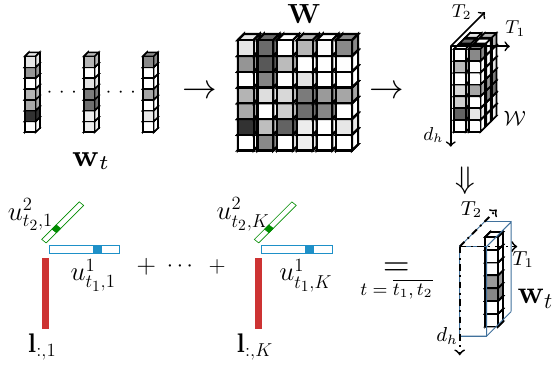}
    \caption{An illustration on our tensorized representations.}
    \label{fig: wt-to-w}
\end{figure}

Assuming $T$ tasks are involved with data $\{\mathbf{x}_{i}^{t}\in\mathbb{R}^{d_{t}},y_{i}^{t}\in\mathbb{R}\}_{i=1}^{m_{t}}$, $T$ sets of parameters $\{\mathbf{w}_t, b^t\}_{t=1}^T$ are thereby required for predictions in MTL. 
Here we focus on homogeneous attributes with  $d_{t} = d$. Thus, the complete weight matrix is   $\mathbf W=[\mathbf w_1; \ldots; \mathbf w_T]$. 
Instead of using a single index for these $T$ tasks, multiple indices for an efficient and structured representation can be considered to construct a higher-order tensor \cite{romera2013multilinear}.
{In this paper,   the weight tensor is constructed as  $\mathcal{W}\in\mathbb{R}^{d_h\times T_1\times\cdots T_N}$ and we factorize it into CP form for the structural relatedness across different tasks, such that:}
\begin{equation}\label{eq:tensor:w:cp}
\mathcal{W}=\sum\nolimits_{k=1}^K \mathbf{l}_{:,k}\circ \mathbf{u}^{1}_{:,k}\circ \cdots\circ \mathbf{u}^{N}_{:,k}=[\![ \mathbf{L},\mathbf{U}^{1},\ldots,\mathbf{U}^{N}]\!],
\end{equation}
where $\mathbf{L}=[\mathbf{l}_{:,1}; \cdots; \mathbf{l}_{:,K}]\in\mathbb{R}^{d_h\times K}$ is the shared {factor} exploiting coupling information across tasks, $\mathbf{U}^n=[\mathbf{u}^n_{:,1}, \ldots, \mathbf{u}^n_{:,K} ]  \in\mathbb{R}^{T_n\times K}$ corresponds to the $n$-th index with $\mathbf{u}^n_{:,k}=[u^n_{1,k},\ldots,u^n_{T_n,k}]^\top$. 
The task-specific coefficient is thus formulated as:
\begin{equation}\label{eq:tensor:w:vector}
\mathbf{w}_t=\sum\nolimits_{k=1}^K \mathbf{l}_{:,k}\cdot {u}^{1}_{t_1,k}  \ldots {u}^{N}_{t_N,k}.
\end{equation} 
Each task  is now spanned by $N$ indices, i.e., $t=\overline{t_1,\ldots,t_N}$ with $t_n=1,\ldots, T_n$, $n=1,\ldots,N$, so that the total number of tasks is  calculated by $T = \prod_{n=1}^N T_n$. Fig. \ref{fig: wt-to-w} gives a graphical illustration for a third-order case.

It is explicit that $\{\mathbf{l}_{:,1},\cdots,\mathbf{l}_{:,K}\}$ learns the coupling information across tasks and is always involved in the prediction for each task.
In contrast, the variation  of $\mathbf{u}_{t_n,:}^n$ affects a certain group  of tasks relating to the index $t_n$. For instance, for $n=1$, $t_1=1$, the updating of $\mathbf{u}_{1,:}^1$ affects tasks in $\{t=\overline{1,\ldots,t_N}|t_l=1,\cdots,T_l,l\neq 1\}$.
In other words, the correlations between tasks can be explored by splitting them into different modes (indices) 
with a high-order tensor, enabling {structural} captures of  dependencies from multiple modes than using a single mode.  In this way, CP rank $K$ indicates the number of latent shared features $\mathbf{l}_{:,k}$ in this representation.
With the imposed low CP rank, the learned coefficients can be more compact in gaining informative modelling.

Then, our {tensorized} LSSVM for  MTL regression, i.e., tLSSVM-MTL, is constructed in the primal form as: 
\begin{eqnarray}\label{eq:our:lsstm:primal}
\underset{\mathbf{L}, \mathbf{U}^{n}, b^{t},e_i^t}{\min } & \frac{C}{2} \sum\limits_{t=1}^T\sum\limits_{i=1}^{m_t}(e_{i}^{t})^{2}+\frac{1}{2}\operatorname{tr} \mathbf{L L}^{\top}+\frac{1}{2}\sum\limits_{n=1}^{N} \operatorname{tr} \mathbf{U}^{n}  {\mathbf{U}^{n}}^{\top} \nonumber\\ 
\text { s.t. } &
( \sum\nolimits_{k=1}^{K} (\mathbf{l}_{:,k}  \cdot {u}_{t_{1},k}^{1}  \ldots {u}_{t_{N},k}^{N}  ))^{\top} \phi(\mathbf{x}_{i}^{t})+b^{t} \\
& =y_{i}^{t}-e_{i}^{t}, \quad t=\overline{t_1,\ldots,t_N}. \nonumber
\end{eqnarray}
{With the constructed tensor and the deployed factorization,
our proposed tLSSVM-MTL successfully extends the existing LSSVMs  to deal with multitasks referenced by multiple indices;} the low CP rank factorization enables to explicitly attain the shared factor $\mathbf{L}$  seeking for common information and these $\mathbf{U}^n$ maintaining task-specific information, which together boosts the overall performance of all tasks.



\subsection{Optimization Algorithm}\label{sec:algorithm}
In
\eqref{eq:our:lsstm:primal}, the product operations  between the shared  $\mathbf{L}$ and the task-specific  $\mathbf{U}^1,\ldots, \mathbf{U}^N$ 
result in nonconvexity, {but can be decoupled by block coordinate descents. We thus design an alternating updating strategy to optimize each factor iteratively, where each subproblem successfully degenerates to be convex by solving a linear system with Lagrangian duality.}

\vspace{0.2cm}
\noindent 1) {\bf{Step $\mathbf{L},b^t,e_i^t$ with fixed $\mathbf{U}^n$}.}
The primal problem with respect to $\mathbf{L},b^t,e_i^t$ is given by
\begin{eqnarray}\label{eq:update:l}
\underset{\mathbf{L}, b^{t},e_i^t}{\min } & \frac{{C}}{2} \sum\nolimits_{t=1}^{T} \sum\nolimits_{i=1}^{m_{t}}\left(e_{i}^{t}\right)^{2}+\frac{1}{2}\operatorname{tr} (\mathbf {L L}^{\top})\nonumber\\
\text { s.~t. } & 
( \sum\nolimits_{k=1}^K \left(\mathbf{l}_{:,k}  \cdot {u}_{t,k}\right))^{\top} \phi(\mathbf{x}_{i}^{t})+b^{t}
=y_{i}^{t}-e_{i}^{t}, \nonumber
\end{eqnarray}
where $u_{t,k}\triangleq {u}_{t_{1},k}^{1}  \cdots {u}_{t_{N},k}^{N}$ for $t=\overline{t_{1},\ldots,t_{N}}, t_{n}=1,\ldots,T_{n}$. 
With dual variables $\alpha_i^t \in \mathbb{R}$ corresponding to each equality constraint, the Lagrangian function is obtained as
 \begin{equation}
\mathcal{L}\left(\mathbf{L}, b^{t},e_i^t\right)=\frac{{C}}{2} \sum\nolimits_{t=1}^{T} \sum\nolimits_{i=1}^{m_{t}}\left(e_{i}^{t}\right)^{2}+\frac{1}{2}\operatorname{tr} (\mathbf{L L}^{\top})
-\sum\nolimits_{t=1}^T\sum\nolimits_{i=1}^{m_t} \alpha_{i}^{t}(  \left(\mathbf{L}\mathbf{u}_t\right)^{\top} \phi(\mathbf{x}_{i}^{t})+b^{t}-y_{i}^{t}+e_{i}^{t}),\nonumber
\end{equation}
with $\mathbf{u}_{t} \triangleq [u_{t,1}, \ldots, u_{t, K}]^{\top}\in\mathbb{R}^{K}$. Then, stationary point conditions are obtained as
\begin{eqnarray}\label{eq:kkt:l}
\frac{\partial \mathcal{L}}{\partial \mathbf{L}}=0 &\Longrightarrow& \mathbf{L}=\sum\nolimits_{t=1}^T \sum\nolimits_{i=1}^{m_t} \alpha_{i}^{t}  \phi(\mathbf{x}_{i}^{t}) \mathbf{u}_{t}^{\top},\nonumber \\
\frac{\partial \mathcal{L}}{\partial \mathbf{b}}=0 &\Longrightarrow & \mathbf{A}^{\top} \bm \alpha =0,  \ \mathbf{b} = \left[b^{1}, \ldots, b^{T}\right]^{\top}, \nonumber\\ 
\frac{\partial \mathcal{L}}{\partial \mathbf{e}}=0 &\Longrightarrow & C \mathbf{e}=\boldsymbol{\alpha}, \nonumber\\ 
\frac{\partial \mathcal{L}}{\partial \boldsymbol{\alpha}}=0 &\Longrightarrow& \Phi {\mathbf{w}}+ \mathbf{A}\mathbf{b}=\mathbf{y}-\mathbf{e}.\nonumber
 \end{eqnarray}
 {where $\mathbf{A}=\text{blockdiag}(\mathbf{1}_{m_1},\cdots,\mathbf{1}_{m_T})\in\mathbb{R}^{m\times T}$, $\mathbf{w}=[(\mathbf{L}\mathbf{u}_1)^{\top},\cdots,(\mathbf{L}\mathbf{u}_T)^{\top}]^{\top}\in\mathbb{R}^{Td_h}$,  the task-specific feature mapping matrix $\Phi^t=[\phi(x_1^t),\ldots,\phi(x_{m_t}^t)]^{\top}\in\mathbb{R}^{m_t\times d_h}$  and
   $\Phi=\text{blockdiag}(\Phi^1,\cdots,\Phi^T)\in\mathbb{R}^{m\times Td_h}$  for all $T$ tasks.  All outputs, regression errors, and dual variables  are denoted as  $\mathbf{y}=[y_1^1,y_2^1,\ldots,y_{m_T}^T]^{\top}\in\mathbb{R}^{m}$, 
 $\mathbf{e}=[e_1^1,e_2^1,\ldots,e_{m_T}^T]^{\top}\in\mathbb{R}^{m}$, and
 ${\boldsymbol{\alpha}}=[\alpha_1^1,\alpha_2^1,\ldots,\alpha_{m_T}^T]^{\top}\in\mathbb{R}^{m}$, respectively.}

By  eliminating $\mathbf{L}$ and $e^t_i$, a linear system is attained as: 
\begin{equation}
    \left[\begin{array}{c|c}\mathbf{0}_{T \times T} & \mathbf{A}^{\top}\\ \hline \mathbf{A} & \mathbf{\mathbf { Q }}+\frac{1}{C} \mathbf{I}_{m\times m}\end{array}\right]\left[\begin{array}{c} \mathbf{b}\\ \boldsymbol{\alpha}\end{array}\right]=\left[\begin{array}{c}\mathbf{0}_T\\ \mathbf{y}\end{array}\right],
    \label{eq: update_L}
\end{equation}
where  
$\mathbf{Q}\in\mathbb{R}^{m\times m}$ is   computed by the components in  tensor $\mathcal{W}$ and the kernel function $k: \mathbb R^d \times \mathbb R^d \mapsto \mathbb R$ {induced by  $\phi(\cdot)$}, such that $\mathbf{Q}(j,j')=\left\langle\mathbf{u}_{t}, \mathbf{u}_{q}\right\rangle  k\left(\mathbf{x}_{i}^{t}, \mathbf{x}_{p}^{q}\right)$, 
$j=\sum_{r=1}^{t-1} m_r +i, j'=\sum_{r=1}^{q-1} m_r +p, i=1,\cdots,m_t, p=1,\cdots,m_q$ with  $i, p$ indexing the  samples in the involved tasks $t$ and $q$, respectively. With the solution of  dual variables (\ref{eq: update_L}), i.e., $\tilde{\boldsymbol{\alpha}}$,
we can get the updated  $\mathbf{L}=\sum_{t=1}^T \sum_{i=1}^{m_t} \tilde{\alpha}_{i}^{t}  \phi(\mathbf{x}_{i}^{t}) \mathbf{u}_{t}^{\top}$. 
\vspace{0.2cm}
\noindent 2) {\bf{Step $\mathbf{U}^{n},b^t,e_i^t$ with fixed $\mathbf{L}$}.}
With fixed $\mathbf L$, we alternate to optimize $\mathbf{U}^{n},b^t,e_i^t$. 
The corresponding primal problem is:
\begin{eqnarray}
\underset{ \mathbf{u}^{n}_{t_n,:}, b^{t}, e_i^t}{\min }  &\frac{{C}}{2} \sum\nolimits_{t\in\mathbb{S}_{t_n}} \sum\nolimits_{i=1}^{m_{t}}\left(e_{i}^{t}\right)^{2}+ \frac{1}{2}\lVert \mathbf{u}_{t_{n},:}^{n}\lVert_{2}^2  \nonumber\\ \nonumber
\text { s.~t. }  & {\mathbf{u}_{t_{n},:}^{n} }^{\top}\mathbf{z}_{i}^{t} +b^{t}=y_{i}^{t}-e_{i}^{t},  \nonumber
\end{eqnarray}
where 
$\mathbf{z}_{i}^{t}$ is  calculated by $\mathbf{L}^{\top}  {\phi(\mathbf{x}_{i}^{t}) } \odot \mathbf{u}_{t_{1},:}^1\odot\cdots \mathbf{u}_{t_{n-1},:}^{n-1}\odot\mathbf{u}_{t_{n+1},:}^{n+1}\odot\cdots \mathbf{u}_{t_{N},:}^N\in\mathbb{R}^K$, the involved tasks $t$ is contained in the index set
$\mathbb{S}_{t_n}=\{\overline{t_{1}, \ldots, t_{N}}| t_{l}=1,\ldots,T_{l}, l=1, \ldots, N, l \neq n\}$ with cardinality
$|\mathbb{S}_{t_n}|=\prod_{l,l\neq n}T_l$.
With dual variables $\boldsymbol{\lambda}_{t_n}$, we have the Lagrangian function:
\begin{equation}
\mathcal{L}\left(\mathbf{u}^{n}_{t_{n},:}, b^{t}, {e}_i^t\right)=  \frac{{C}}{2} \sum\nolimits_{t\in\mathbb{S}_{t_n}} \sum\nolimits_{i=1}^{m_{t}}\left(e_{i}^{t}\right)^{2}+ \frac{1}{2}\lVert \mathbf{u}_{t_{n},:}^{n}\lVert_{2} ^2\\
-\sum\nolimits_{t\in\mathbb{S}_{t_n}} \sum\nolimits_{i=1}^{m_{t}}\lambda_i^t\left( \left(  {\mathbf{u}_{t_{n},:}^{n} }^{\top}\mathbf{z}_{i}^{t} +b^{t}\right)-y_{i}^{t}+e_{i}^{t} \right),  \nonumber
\end{equation}
where $\boldsymbol{\lambda}_{t_n} = \{\lambda^t_{i}|t\in \mathbb S_{t_n}, i=1, \ldots, m_t\}\in \mathbb{R}^{M_{t_n}}$ corresponds to  the involved constraints  in optimizing $\mathbf{u}^{n}_{t_{n},:}$.

Similarly, by deriving  the stationary  conditions and eliminating $\mathbf{u}_{t_n,:}^n$ and $e^t_i$ therein, we get the  linear system: 
\begin{equation}
    \left[\begin{array}{c|c}
    \mathbf{0}_{\lvert \mathbb{S}_{t_n}\lvert \times \lvert \mathbb{S}_{t_n}\lvert} & \mathbf{A}_{t_n}^{\top}\\ 
      \hline \mathbf{A}_{t_n} & \mathbf{\mathbf { Q_{t_n} }}+\frac{1}{C} \mathbf{I}_{M_{t_n}}\end{array}\right]\left[\begin{array}{c}\mathbf{b}_{t_n}\\
    \boldsymbol{\lambda}_{t_n}\end{array}\right]=\left[\begin{array}{c}\mathbf{0}_{\lvert \mathbb{S}_{t_n}\lvert}\\ \mathbf{y}_{t_n}\end{array}\right],
    \label{eq: update_U}
\end{equation}
where $\mathbf{A}_{t_n}=\text{blockdiag}(\mathbf{1}_{m_t})\in\mathbb{R}^{M_{t_n}\times \lvert \mathbb{S}_{t_n}\lvert}$ with $t\in\mathbb{S}_{t_n}$, and ${\mathbf{y}}_{t_n}$,  ${\boldsymbol{\alpha}}_{t_n}, \mathbf{b}_{t_n} \in \mathbb R^{M_{t_n}}$ are vectors collecting  $y_i^t$, ${\alpha}_i^t$, and $b_i^t$ involved in the equality constraints, respectively.
Here, the  matrix 
$\mathbf{Q}_{t_n}\in\mathbb{R}^{M_{t_n}\times M_{t_n}}$ is computed by $\mathbf{Q}_{t_n}(j,j')=\left\langle\mathbf{z}^{t}_i, \mathbf{z}_{p}^q\right\rangle $, 
where $ t, q\in\mathbb{S}_{t_n}, i=1, \ldots, m_{t},p=1,\cdots,m_q$.

{The proposed alternating algorithm gives the final solutions  after  convergence. In this paper, we set the convergence condition for factors $\mathbf{U}^{n}$, such that $\sum_n \lVert\mathbf{U}^{n}_{k+1}-\mathbf{U}^{n}_{k}\lVert_{\operatorname{F}}^2/\lVert\mathbf{U}^{n}_{k}\lVert_{\operatorname{F}}^2<10^{-3}$.}
After optimization,  the  prediction  for any given input {$\mathbf x$ of the $t$-th task} is obtained either with 
\begin{itemize}
    \item  the expression 1) using explicit feature map $\phi(\cdot)$:
\begin{equation}
f_t(\mathbf{x})=\left(\mathbf{L}\mathbf{u}_t\right)^{\top} \phi(\mathbf{x})+b^{t} 
\end{equation}

\item the expression 2)  using  kernel function $ k(\cdot, \cdot)$:  
\begin{equation}
f_t(\mathbf{x})=\sum\nolimits_{p=1}^{m_q}\sum\nolimits_{q=1}^{T} \lambda_p^q k( \mathbf x, \mathbf x_p^q)\langle\mathbf{u}_t,\mathbf{u}_q\rangle +b^t.
\end{equation}
\end{itemize}
{Note that expression 1) is the primal representation, while 
expression 2) is not strictly the dual representation, due to the existence of parameters $\mathbf{u}_t, \mathbf{u}_q$ in the primal. This is because the optimization algorithm alternates to update different factors of the tensor and the resulting Lagrangian dual forms correspond to each subproblem during iterations, not to the original nonconvex problem \eqref{eq:our:lsstm:primal}.  Nonetheless, the  problem can be efficiently resolved by sets of linear systems, and both expressions 1) and 2) consider correlations across tasks and task-specific information.}




\section{Numerical Experiments}
\label{sec: experiment}
We evaluate the performance of the proposed method on both synthetic and real-world data. Root mean square error (RMSE), $Q^2$
, and the correlation of the predicted $\hat{\mathbf{y}}$ and the ground-truth $\mathbf{y}$ are measured, where
$Q^2$ is defined as $1-\lVert \mathbf{y}-\hat{\mathbf{y}}\lVert_{\operatorname{F}}^2/\lVert \mathbf{y}\lVert_{\operatorname{F}}^2$ and 
each iterative method is repeated 10 times for an average. Except for RMSE, a higher metric value indicates a better result. 
There are three hyperparameters to be tuned in our tLSSVM-MTL, i.e., $K$, $C$, and  the kernel function, {and the  hyperparameters in the compared methods are also tuned, where $5$-fold cross-validation is used.}

\vspace{0.2cm}
\noindent \textbf{1) Simulated data}

The simulated dataset is generated as: 1)  the coefficient tensor via the CP form $\mathcal{W}=[\![\mathbf{L}, \mathbf{U}^{1},\cdots,\mathbf{U}^{N}]\!]$, where each entry  is randomly generated  from $\mathcal{N}(0,1)$; 2)  $\mathbf{x}^t_i$,  $b^t$ and noise ${e}^t_i$  from distribution $\mathcal{N}(0,1)$; 3) the  response  $\mathbf{y}^t=\mathbf{y}^t+\sigma \mathbf{e}^t$ consisting of  $\mathbf{y}^t=\mathbf{X}^t \sum_{k=1}^K \mathbf{l}_k \cdot u^1_{t_1,k}\ldots u^N_{t_N,k}+{b}^t\mathbf 1_{m_t}$ and  $\mathbf{e}^t$ given by the signal-to-noise ratio (SNR).
We set $d=100$, $N=3$, $T_1=3$, $T_2=4$, $T_3=5$ with  $T=60$ tasks,  $K=3$, and 60 training samples  and  20 test samples for each task.


\begin{figure}[ht!]
\centering   
         \includegraphics[width=0.55\textwidth]{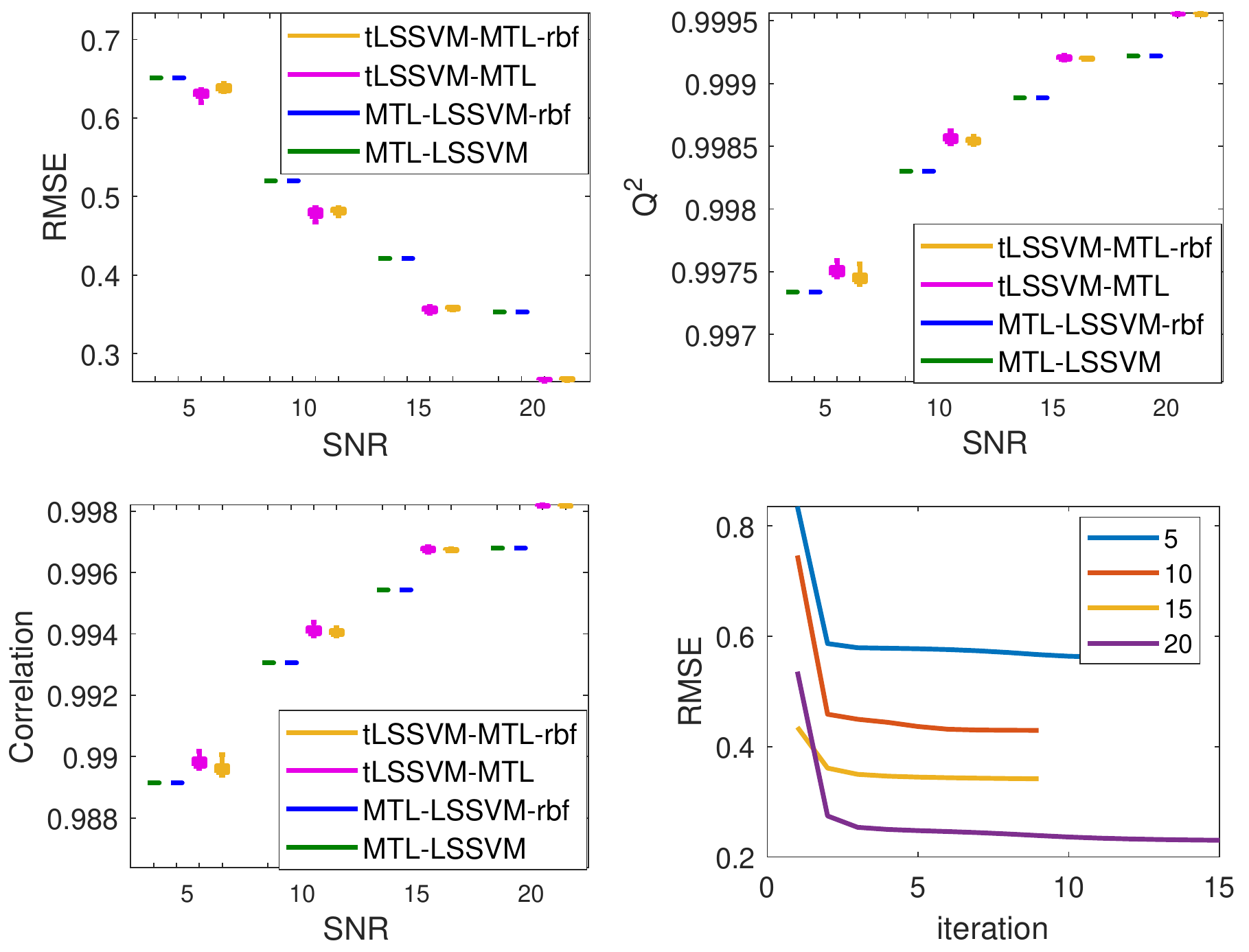}
    \caption{Performance on simulated data with different SNRs. }
    \label{fig:com-sim}
\end{figure}

{This experiment mainly aims to validate the efficacy of our tensorized tLSSVM-MTL and optimization results of the proposed algorithm; thus, the  MTL-LSSVM counterpart is compared.  Fig. \ref{fig:com-sim} presents the performance evaluations on simulated data with different SNR levels, showing that the proposed tLSSVM-MTL consistently provides more accurate predictions on varied SNRs, and its advantage is slightly better with larger SNRs. Additionally, we plot the RMSE during the iterative updates in our method, where RMSE sharply decreases and then converges to a small error. The results of this experiment verify the effectiveness of the proposed method.}

\vspace{0.2cm}
\noindent \textbf{2) Real-world Data}

Three  datasets for MTL are employed: Restaurant \& Consumer \cite{vargas2011effects}, Student performance \footnote{\url{ https://archive.ics.uci.edu/ml/datasets/Student+Performance}}, and  Comprehensive Climate (CCDS). 
 Restaurant \& Consumer Dataset contains the rating scores of $138$ consumers to different restaurants in  3 aspects, leading to $138\times 3$ regression tasks.   Student performance Dataset contains student grades in 3 periods and {other attributes like sex, and age}, where we build $3 \times 2$ regression tasks by {separating the data according to sex and grade period}.  Comprehensive Climate Dataset (CCDS) gives monthly climate records of $17$ variables in North America from 1990 to 2001 \cite{lozano2009spatio}, where we select $5$ locations and construct $5 \times 17$ regression tasks.
 MTL-LSSVM \cite{xu2014multi} and two tensor-based  methods, i.e., Convex and Nonconvex Multilinear MTL (MLMTL-C and MLMTL-NC) \cite{romera2013multilinear}, are  compared.

\begin{table}[ht!]
    
    \centering
     \scalebox{0.94}{
   \begin{tabular}{ |c|c|c|c|c| }
 \hline
 \multicolumn{5}{|c|}{ Restaurant \& Consumer}\\
\hline
Metric & RMSE & $Q^2$ & Correlation &CPU Time\\
\hline
MTL-LSSVM&0.65&41.83\%&62.54\%&0.45\\
MTL-LSSVM-rbf&0.65&41.90\%&62.55\%&0.51\\
MLMTL-C&0.65& 40.42 \%&61.31\%& \textbf{0.45}\\
MLMTL-NC&0.74 &18.61\%&56.12\%&41.10\\
tLSSVM-MTL&0.61&45.41\%&67.03\%&22.86\\
tLSSVM-MTL-rbf&\textbf{0.59}&\textbf{49.13}\%&\textbf{69.54}\%&19.36\\
\hline
\end{tabular}}
    \scalebox{0.95}{
   \begin{tabular}{ |c|c|c|c|c| }
   \hline
    \multicolumn{5}{|c|}{ Student Performance}\\
\hline
Metric & RMSE & $Q^2$ & Correlation &CPU Time\\
\hline
MTL-LSSVM & 2.99&93.55\%&44.66\%& \textbf{0.03}\\
MTL-LSSVM-rbf& 2.49&95.56\%&67.49\%&0.04\\
MLMTL-C& 3.11& 93.03\% &36.45\%& 3.21\\
MLMTL-NC&3.34&91.96\%&21.51\%&19.10\\
tLSSVM-MTL&2.99&93.54\%&45.79\%&0.72\\
tLSSVM-MTL-rbf&\textbf{2.44}&\textbf{95.73}\%&\textbf{68.59}\%&0.41 \\
\hline
\end{tabular}}

\scalebox{0.95}{
   \begin{tabular}{ |c|c|c|c|c| }
   \hline
    \multicolumn{5}{|c|}{ CCDS}\\
\hline
Metric & RMSE & $Q^2$ & Correlation &CPU Time\\
\hline
MTL-LSSVM & 0.79&29.71\%&55.50\%& \textbf{1.08}\\
MTL-LSSVM-rbf& {0.70}&{46.70}\%&{68.36}\%&1.50\\
MLMTL-C& 0.76&34.56\% &58.79\%& 5.31\\
MLMTL-NC&0.83&24.04\%&50.02\%&29.44\\
tLSSVM-MTL&0.78&32.64\%&58.03\%&24.07\\
tLSSVM-MTL-rbf&\textbf{0.65}&\textbf{54.50}\%&\textbf{74.49}\%&22.01 \\
\hline
\end{tabular}}
    \vspace{0.4cm}
    \caption{Performance comparison on real-world datasets.}
        \vspace{-0.4cm}
    \label{tab:real_exp}
\end{table}


 Table \ref{tab:real_exp} presents the prediction results by MTL-LSSVM, MLMTL-C, MLMTL-NC, and the proposed tLSSVM-MTL with both linear and RBF kernels, where the best results are in bold. The results show that our proposed method substantially improves the prediction accuracy in terms of all considered metrics. Our advantages appear  more  prominent for Restaurant \& Consumer and CCDS datasets with RBF kernels, particularly on $Q^2$ and Correlation metrics which achieve significant improvements. In fact, these two datasets contain larger numbers of tasks, i.e., $T=414$ and $T=35$, and the used multiple indices are endowed with specific meanings in prior to their real-world applications, thereby enabling our  model to well  learn the underlying structural information.
 
 In Table \ref{tab:real_exp}, we also compare the CPU time. We can see that the existing matrix-based MTL-LSSVM and MLMTL-C run faster, due to their convexity benefiting a simple optimization. When comparing with the nonconvex tensor-based MLMTL-NC, our method is more efficient, particularly for the Student Performance dataset, still showing the promising potentials of our tensorized model and the designed iterative updates. Nevertheless, more efficient computations can be expected with further investigations.  

\vspace{-0.3cm}
\section{Conclusion}
\label{sec: conclusion}
\vspace{-0.2cm}

In this paper, we proposed a novel method for MTL regression, {which can be regarded as a tensorized generalization and also  a multimodal extension of multitask LSSVMs.} 
The proposed method considers multitasks with different indices in the constructed coefficient tensor, {which is factorized with low CP rank into a common factor and task-specific factors. In the proposed method, both multilinear and nonlinearity can be flexibly modelled either through  feature mappings  or kernel functions.
In optimization, an alternating  strategy is derived to update these factors by solving linear programming subproblems with Lagrangian duality. Experimental results on simulated and real-world data show our great potentials}  over the compared relevant methods. In future, different tensorization techniques and faster computations are promising to be extended to wider ranges of tasks.


\bibliographystyle{plain}
\bibliography{arxiv}

\end{document}